%% file: jsen.tex
\documentclass[journal,twoside,web]{ieeecolor}
\usepackage{jsen}
\usepackage{cite}
\usepackage{amsmath,amssymb,amsfonts}
\usepackage{algorithmic}
\usepackage{graphicx}
\usepackage{textcomp}
\usepackage{wrapfig}
\usepackage[hidelinks]{hyperref}
\usepackage{cleveref}
\usepackage{booktabs}
\usepackage{xcolor}
\usepackage{siunitx}
\usepackage[acronym,shortcuts]{glossaries}

\let\labelindent\relax
\usepackage[inline]{enumitem}

\DeclareSIUnit{\fps}{FPS}

\def\BibTeX{{\rm B\kern-.05em{\sc i\kern-.025em b}\kern-.08em
    T\kern-.1667em\lower.7ex\hbox{E}\kern-.125emX}}
\definecolor{abstractbg}{rgb}{0.89804,0.94510,0.83137}
\usepackage{eso-pic}
\usepackage{url}
\AddToShipoutPictureBG*{
  \AtPageUpperLeft{%
    \put(0,-40){\raisebox{15pt}{\makebox[\paperwidth]{\begin{minipage}{21cm}\centering
      \textcolor{gray}{This article has been accepted for publication in the IEEE Sensors Journal (JSEN). \\ DOI: 10.1109/JSEN.2025.3644788} 
    \end{minipage}}}}%
  }
  \AtPageLowerLeft{%
    \raisebox{19pt}{\makebox[\paperwidth]{\begin{minipage}{21cm}\centering
      \textcolor{gray}{ \copyright 2026 IEEE.  Personal use of this material is permitted.  Permission from IEEE must be obtained for all other uses, in any current or future media, including reprinting/republishing this material for advertising or promotional purposes, creating new collective works, for resale or redistribution to servers or lists, or reuse of any copyrighted component of this work in other works.
      }
    \end{minipage}}}%
  }
}

\begin{document}
\include{acr.tex}
\title{LEVIO: Lightweight Embedded Visual Inertial Odometry for Resource-Constrained Devices}
\author{Jonas Kühne,~\IEEEmembership{Member,~IEEE}, 
Christian Vogt,~\IEEEmembership{Member,~IEEE},\\ 
Michele Magno,~\IEEEmembership{Senior Member,~IEEE},
and Luca Benini,~\IEEEmembership{Fellow,~IEEE}
\thanks{Jonas Kühne is with the Integrated Systems Laboratory and the Center for Project-Based Learning, ETH Zurich, 8092 Zurich, Switzerland (e-mail: kuehnej@ethz.ch).}
\thanks{Christian Vogt is with the Center for Project-Based Learning, ETH Zurich, 8092 Zurich, Switzerland (e-mail: christian.vogt@pbl.ee.ethz.ch).}
\thanks{Michele Magno is with the Center for Project-Based Learning, ETH Zurich, 8092 Zurich, Switzerland (e-mail: michele.magno@pbl.ee.ethz.ch).}
\thanks{Luca Benini is with the Integrated Systems Laboratory, ETH Zurich, 8092 Zurich, Switzerland, and also with the Department of Electrical, Electronic and Information Engineering, University of Bologna, 40136 Bologna, Italy (e-mail: luca.benini@unibo.it).}}

\IEEEtitleabstractindextext{%
\fcolorbox{abstractbg}{abstractbg}{%
\begin{minipage}{\textwidth}%
\begin{wrapfigure}[13]{r}{3in}%
\includegraphics{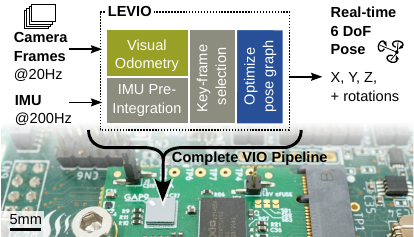}%
\end{wrapfigure}%
\begin{abstract}
Accurate, infrastructure-less sensor systems for motion tracking are essential for mobile robotics and augmented reality (AR) applications. The most popular state-of-the-art visual-inertial odometry (VIO) systems, however, are too computationally demanding for resource-constrained hardware, such as
micro-drones and smart glasses. This work presents LEVIO, a fully featured VIO pipeline optimized for ultra-low-power compute platforms, allowing six-degrees-of-freedom (DoF) real-time sensing. LEVIO incorporates established VIO components such as Oriented FAST and Rotated BRIEF (ORB) feature tracking and bundle adjustment, while emphasizing a computationally efficient architecture with parallelization and low memory usage to suit embedded microcontrollers and low-power systems-on-chip (SoCs). The paper proposes and details the algorithmic design choices and the hardware-software co-optimization approach, and presents real-time performance on resource-constrained hardware. LEVIO is validated on a parallel-processing ultra-low-power RISC-V SoC, achieving 20 FPS while consuming less than 100\,mW, and benchmarked against public VIO datasets, offering a compelling balance between efficiency and accuracy. To facilitate reproducibility and adoption, the complete implementation is released as open-source.

\end{abstract}

\end{minipage}}}

\maketitle

\section{Introduction}
\label{sec:levio_introduction}

In robotic and augmented reality tasks, tracking the device's movement relative to its surroundings is essential \cite{rana2024enhanced}. While infrastructure-based indoor position sensing can either be achieved with cost intensive high performance setups (e.g., motion tracking \cite{sesyuk2022survey}), or at a lower accuracy and cost (e.g., radio localization with Wi-Fi \cite{rana2024enhanced,sun2023multiuser} or ultra-wide-band anchors \cite{laadung2021novel}) the resulting infrastructure restricts the tracking to a defined area and generates additional recurring costs of infrastructure maintenance. To enable truly autonomous operation in robotics, both indoors and outdoors, and full freedom of movement for augmented reality tasks, future movement tracking needs to be infrastructure-less and accurate\,\cite{savic2022constrained}.

Infrastructure-less position-tracking systems are an established research field, mostly focusing on camera sensors in combination with \gls{vo} algorithms \cite{macario2022comprehensive}. This is a practical approach \cite{xu2023pvi}, thanks to the wide availability and affordability of small and low-power camera modules. \Gls{vo} algorithms are often augmented with \gls{imu} readings, forming a \gls{vio} system to achieve higher tracking accuracy, with \gls{imu} and \gls{vo} complementing each other. Depending on the application, \gls{vio} can be extended to Visual Inertial \gls{slam}, enabling the detection of previously visited places as well as the optimization and retroactive correction of the predicted trajectory through place recognition \cite{macario2022comprehensive,merzlyakov2021comparison}. 

While both \gls{vio} and Visual Inertial \gls{slam} algorithms have been deployed in a multitude of applications in robotics, automotive, and mixed reality, the current state-of-the-art algorithms are computationally expensive \cite{zeng2023fast}. They typically require powerful computing platforms with power requirements ranging from watts to tens of watts \cite{xu2023pvi}, to process camera streams with \SIrange{20}{50}{\fps} \cite{bloesch2015robust}. Since those computing platforms and the corresponding battery packs to power them can weigh hundreds of grams to kilograms, the application of state-of-the-art \gls{vio} in micro- and nano-drones, as well as in augmented reality glasses, is limited or prevented up to now \cite{zeng2023fast}, due to strict payload limitations.

Although modern \gls{vio} algorithms are not available on small-scale devices due to the challenges of computing at low latency on ultra-low-power processors, previous works reduced the full \gls{vio} problem to simpler formulations capable of operating with less compute power. Among the least power-demanding implementations is the fully integrated \gls{asic} Navion \cite{suleiman2019navion}, consuming down to \SI{2}{\milli\watt} at 20 frames per second. Another example of highly constrained algorithms is the reduction of \gls{vio} to planar localization only \cite{honegger2013open, kuhne2022parallelizing}. Further increasing power budget and relaxing algorithmic constraints, research utilizing \glspl{fpga} or even Raspberry Pis has been shown to run \gls{vio} pipelines by using specialized hardware for optical flow extraction on-camera \cite{kuhne2024low}, or \gls{fpga}-based hardware accelerators \cite{gu2022real}.

Following up on these previous works to reduce the energy footprint of \gls{vio} towards accurate, infrastructure-less, and low power movement tracking, this work bridges the gap between the simplified algorithms typically used on constrained devices and accurate state-of-the-art algorithms, by proposing a fully featured and lightweight six \gls{dof} \gls{vio} pipeline optimized to run on ultra-low power compute platforms. Additionally, this new \gls{vio} pipeline, termed \gls{levio}, is fully implemented in C code to be deployable on \gls{cots} hardware and open-sourced.

The main contributions of \gls{levio} are:
\begin{enumerate}
    \item \textbf{Efficient \gls{vio} Pipeline:} Through design space exploration, we propose a six \gls{dof} \gls{vio} pipeline termed \gls{levio}, which is lightweight and accurate. The \gls{levio} pipeline consists of established \gls{vio} components and algorithms like \gls{orb} features and \gls{ba}, while focusing on a slim, computationally efficient design, able to process image streams at \SI{20}{\fps}.
    \item \textbf{Hardware-Software Codesign:} The proposed \gls{levio} pipeline is further tailored towards embedded microcontrollers and \glspl{soc}, with strict memory limitations and limited computational power in mind. Therefore, we highlight the relevant design choices and optimization for parallelization and memory management, resulting in a highly parallelizable algorithm to leverage modern multicore platforms and operate with a low memory footprint (below \SI{256}{\kilo \byte} for \gls{vo} without optimization and below \SI{1}{\mega \byte} for the whole pipeline). 
    \item \textbf{On-Hardware Validation:} As proof of concept, the proposed pipeline is validated in the form of an implementation for an existing ultra-low power RISC-V \gls{soc} with parallel processing capabilities, and thoroughly benchmarked against open-source \gls{vio} datasets for comparison with other literature. For the validation, the commercially available GAP9 \gls{soc} by GreenWaves Technologies has been chosen as a suitable platform.
    \item \textbf{Open-Source:} To support further advancements in lightweight VIO, we open-source the full implementation of \gls{levio} here: \url{https://github.com/ETH-PBL/levio}
\end{enumerate}

This work is organized as follows:
\Cref{sec:levio_related_work} presents and reviews related work on influential \gls{vio} pipelines and pipelines targeting resource-constrained systems. \Cref{sec:levio_methodology} details the \gls{levio} pipeline and the necessary embedded optimizations to target a low-power \gls{soc}. \Cref{sec:levio_evaluation,sec:levio_results} cover the evaluation setup and experimental results. \Cref{sec:levio_discussion} discusses the implications and limitations of the presented approach. \Cref{sec:levio_conclusion} concludes this work.

\section{Related Work}
\label{sec:levio_related_work}

In this section, we present the previous work related to our \gls{vio} approach. Both \gls{vio} and Visual Inertial \gls{slam} have been extensively studied; therefore, we limit the related work section to the most influential works as well as to the works on which we base our \gls{levio} pipeline. Since our work aims to bridge the GAP between accurate and powerful approaches, as well as approaches targeted at resource-constrained devices, we introduce the corresponding work in separate subsections.

\begin{table*}[t]
\centering
\caption{Comparison of low-power \gls{vo} and \gls{vio} implementations across embedded platforms. The systems are ordered by increasing power requirements. The divider line further separates systems running either on bare-metal or on an embedded \gls{os} (above the divider line) from those systems utilizing a fully featured Linux \gls{os} (below the divider line).}
\label{tab:embedded_vio_refs}
\renewcommand{\arraystretch}{1.2}
\begin{tabular}{@{}lccccc@{}}
\toprule
\textbf{System / Work} 
& \textbf{Year} 
& \textbf{Type} 
& \textbf{Power $\downarrow$} 
& \textbf{Frame Rate $\uparrow$} 
& \textbf{Processor / Platform}  \\
\midrule

Navion \cite{suleiman2019navion}
& 2019
& Monocular VIO 
& \SI{2}{\milli\watt} 
& \SI{20}{\fps} 
& Custom ASIC \\

Parallelized PX4FLOW \cite{kuhne2022parallelizing}
& 2022
& 2D Optical Flow 
& \SI{25}{\milli\watt} 
& \SI{500}{\fps} 
& GreenWaves GAP8 SoC  \\

PX4FLOW \cite{honegger2013open}
& 2013 
& 2D Optical Flow 
& \SI{<100}{\milli\watt} 
& \SI{250}{\fps} 
& ARM Cortex-M4F \\

PicoVO \cite{he2021picovo}
& 2021
& Monocular VO 
& \SI{310}{\milli\watt} 
& \SI{33}{\fps} 
& ARM Cortex-M7  \\

\midrule

Bahnam et al. \cite{bahnam2021stereo}
& 2021
& Stereo VIO (MSCKF) 
& Not stated 
& Not stated 
& RPi Zero \\

Gu et al. \cite{gu2022real}
& 2022
& VO (Hybrid FPGA+CPU) 
& \SI{2.6}{\watt} 
& \SI{20}{\fps} 
& Zynq Ultrascale+ ZU3EG \\

Kühne et al. \cite{kuhne2024low}
& 2024 
& VIO 
& \SI{3.8}{\watt} 
& \SI{50}{\fps} 
& RPi CM4 + On-Sensor Acceleration \\

BotVIO \cite{wei2025botvio}
& 2025
& Monocular VIO (Transformer-Based) 
& \SI{5.4}{\watt} 
& \SI{57.8}{\fps} 
& NVIDIA Jetson Xavier NX \\

\bottomrule
\end{tabular}
\end{table*}

\subsection{Influential Visual Inertial Odometry Works}
\label{sec:related_work_influential_vio}

Among the active research areas of \gls{vo}, \gls{vio}, and \gls{vislam}, which are frameworks for estimating motion and building a map of the environment, a wide variety of algorithms have been proposed, each with different trade-offs and design philosophies. These systems can be classified along several key dimensions:

\begin{itemize}
\item \textbf{Direct vs. Indirect vs. Hybrid Tracking}: This refers to how visual information is processed. Indirect (or feature-based) methods extract and match visual features (e.g., corners, descriptors), whereas direct methods work directly on raw pixel intensities by minimizing the photometric error. Hybrid methods combine aspects of both.
\item \textbf{Filtering vs. Optimization}: Filtering approaches (e.g., \gls{ekf}) process data sequentially and recursively, while optimization-based approaches (e.g., \gls{ba}) refine state estimates over a window of data or keyframes using non-linear optimization.
\item \textbf{Loop Closure}: Indicates whether a system includes the ability to recognize and correct for revisiting previously seen locations to reduce long-term drift.
\item \textbf{Coupling Strategy}: Describes how the camera and inertial measurements are fused. \textit{Tightly coupled} systems jointly estimate motion using both modalities in a unified framework, while \textit{loosely coupled} systems process them separately and fuse results at a higher level.
\item \textbf{Sensor Setup}: Encompasses the use of monocular, stereo, or RGB-D cameras, which affects scale and depth observability, initialization, and robustness.
\end{itemize}

These dimensions greatly affect a system's accuracy, robustness, real-time capability, and suitability for resource-constrained platforms. We subsequently describe several influential works in the field of \gls{vio} and \gls{vislam}, using these dimensions as a framework for comparison. One key similarity among these algorithms is their computational complexity, which typically requires desktop computers for real-time position estimation.

One of the earliest and most well-known indirect, filtering-based VIO systems is MSCKF \cite{mourikis2007multi}, proposed in 2007. It adopts a tightly coupled design, using an EKF to fuse tracked visual SIFT \cite{lowe1999object} features with IMU measurements. MSCKF does not support loop closure and relies on a monocular camera setup. At the time of publication, it was capable of operating at \SI{14}{\hertz} on a single-core \SI{2}{\giga\hertz} desktop processor, demonstrating real-time feasibility.

ROVIO \cite{bloesch2015robust} is a subsequent evolution within the same class of systems (indirect, filtering-based, tightly coupled), but it adopts a more computationally efficient approach. Instead of detecting and matching features, it tracks small image patches (i.e., pixel templates) directly within the \gls{ekf} framework, thus resembling a hybrid method. ROVIO is designed for monocular cameras and omits loop closure. It can process around 50 tracked patches in approximately \SI{30}{\milli\second} per frame on a single core of an Intel i7-2760QM.

To improve long-term accuracy, several systems implement loop closure in addition to VIO. A prominent example is OKVIS (2015) \cite{leutenegger2015keyframe}, a tightly coupled, indirect, optimization-based \gls{vislam} system. OKVIS uses a stereo or monocular camera setup, tracks Harris \cite{harris1988combined} corners, and describes them using BRISK descriptors \cite{leutenegger2011brisk}. The system employs a sliding-window approach for local \gls{ba} and includes loop closure for global consistency. It achieves real-time performance on a quad-core Intel Core i7 at \SI{2.2}{\giga\hertz}.

In contrast to the works presented above, VI-DSO \cite{von2018direct} represents a shift toward direct, optimization-based methods. It tracks camera motion by minimizing the photometric error across frames without relying on keypoint extraction or descriptors. The system uses a tightly coupled fusion of inertial and visual data within a non-linear optimization framework. VI-DSO supports monocular input and does not perform loop closure. The absence of feature extraction improves runtime performance and robustness in low-texture environments.

VINS-Mono \cite{qin2018vins} is another indirect, optimization-based, tightly coupled \gls{vislam} system. It employs KLT optical flow \cite{lucas1981iterative} to track features across frames and maintains a sliding window for non-linear optimization. VINS-Mono supports monocular cameras and includes a robust loop closure and relocalization mechanism via pose-graph optimization. It was demonstrated to run in real-time on an Intel Core i7-5500U.

Concluding the enumeration of influential \gls{vio} works, ORB-SLAM3 \cite{campos2021orb} is one of the most complete and flexible indirect, optimization-based \gls{vislam} systems. It supports monocular, stereo, and RGB-D inputs, with optional integration of IMU data. ORB-SLAM3 performs full \gls{ba} and includes robust loop closure, relocalization, and map reuse. It uses \gls{orb} features \cite{rublee2011orb} for tracking and mapping, which provide a good trade-off between speed and robustness. Thanks to its modularity and performance, it achieves real-time operation on an Intel Core i7-7700 (\SI{3.6}{\giga\hertz}) and has become a widely used baseline in the community.

A comprehensive overview and comparison of these and other state-of-the-art systems is provided by Macario et al. \cite{macario2022comprehensive}, including a taxonomy and benchmarking results across datasets and hardware platforms.

To achieve an accurate six \gls{dof} \gls{vio} pipeline, we draw inspiration from current state-of-the-art \gls{vio} pipelines like VINS-Fusion \cite{qin2018vins} and ORB-SLAM3 \cite{campos2021orb}, and extract the technological advances that led to the most significant performance improvements. Given the strict hardware limitations in compute and memory on an ultra-low-power \gls{soc} like the targeted GAP9 \gls{soc}, we aim to implement the algorithms and pipeline segments that yield high accuracy improvements at moderate computational requirements, while leaving out expensive computations that are used in the top-performing \gls{vio} pipelines.

\subsection{VIO for Resource-Constrained Devices}
Among resource-constrained devices, we review the broad range of \gls{vio} implementations from microcontrollers in the \si{\milli\watt}-range up to single-board computers in the single-digit \si{\watt}-range and summarize the key findings in \Cref{tab:embedded_vio_refs}.
\gls{vio} implementations for microcontrollers typically utilize simplified approaches to achieve low energy consumption. One prime example is PX4FLOW \cite{honegger2013open}. The algorithm is restricted to 2D planar motion and calculates \gls{of} by matching image patches between frames. This reduces computational effort with respect to more complex feature detectors like \gls{orb} \cite{mur2017orb}, but limits the speed at which features can move between two consecutive camera frames. Due to these optimizations, PX4FLOW achieves up to \SI{250}{\hertz} update rates while running on a Cortex M4F processor. An extension of this principle is presented by \cite{kuhne2022parallelizing} on a state-of-the-art \gls{soc} GAP8, improving the PX4FLOW algorithm to run at up to \SI{500}{\hertz}, while consuming only \SI{25}{\milli\watt}.

Increasing complexity to 3D tracking, PicoVO \cite{he2021picovo} presents the feasibility of running \gls{vo}-only on an ARM Cortex M7 with up to \SI{33}{\fps} at \SI{310}{\milli\watt}.

The system called Navion \cite{suleiman2019navion} further enhances the algorithmic capabilities to a full six \gls{dof} \gls{vio} system. This work consumes an average of \SI{2}{\milli\watt} while processing \SI{20}{\fps} thanks to its fully custom-designed \gls{asic} chip.

Similarly to the custom \gls{asic} implementation, when allowing more capable algorithms and compute platforms, one option to still reduce power consumption is to offload computationally expensive tasks to dedicated hardware. In this domain \cite{gu2022real} presents an \gls{fpga} \gls{soc} approach based on an AMD Zynq UltraScale+ ZU3EG, where the compute-intensive Harris corner detection is implemented on the programmable logic part, while the rest of the pipeline resides in the programmable system part of the chip. Overall, the system power consumption is \SI{2.6}{\watt} with a processing time for the frontend of \SI{28.7}{\milli\second}, allowing \SI{20}{\fps}. A similar idea of offloading computation is presented in \cite{kuhne2024low}, which integrates a camera with on-sensor processing capabilities. The camera pre-processes images and outputs the calculated optical flow, which is being used by the VINS-Mono \cite{qin2018vins} pipeline to estimate \gls{vio}. The overall system can process up to \SI{50}{\fps} on a Raspberry Pi Compute Module 4, consuming \SI{3.8}{\watt}. Similarly, an extended algorithm also containing stereo matching (MSCKF) was optimized for an RPi Zero compute module in \cite{bahnam2021stereo}.

Notably, the presented approaches all rely on classical image processing. While machine learning approaches exist, the most recent and optimized designs, like BotVIO \cite{wei2025botvio}, still require a significant amount of power and memory to achieve competitive accuracy. While BotVIO is able to run on a NVIDIA Jetson NX, where only two of the six CPU cores are utilized, it relies on the integrated GPU cores to achieve its top framerate of \SI{57.8}{\fps} at \SI{5.4}{\watt}.

This work advances the state of the art, proposing an efficient six \gls{dof} \gls{vio} pipeline that can be deployed on a low-power \gls{soc} using less than \SI{100}{\milli \watt}, while reaching \SI{20}{\fps}.

\section{Methodology}
\label{sec:levio_methodology}
The \gls{levio} system architecture is guided by a constraint-driven design approach, where the inherent limitations of the targeted ultra-low-power GAP9 \gls{soc}, such as its constrained memory and multicore cluster architecture, led to key innovations within the \gls{vio} pipeline. Unlike \gls{vio} systems designed for desktop platforms, which prioritize accuracy and feature density, \gls{levio}'s design choices were made to ensure real-time performance at sub-100 mW power consumption. In the following subsections, we integrate the description of the \gls{vio} pipeline with a detailed explanation of how these hardware constraints directly shaped our methodology and optimization strategies.

To efficiently explore the design space, we implemented a golden model of \gls{levio} in Python. This model was used to validate and benchmark the \gls{levio} pipeline in terms of accuracy improvements and robustness towards changing parameter configurations. This allows us to design and optimize the \gls{levio} system, validating the performance gain of possible pipeline segments before incorporating them into the final design. The Python golden model is used for algorithmic design-space exploration and parameter sweeps.

Once we obtained the final pipeline architecture (\cref{sec:levio_pipeline_description}), we further tailored it towards ultra-low-power multicore \glspl{soc} and microcontrollers. The modifications required to deploy the pipeline on a \gls{soc} with less than \SI{1}{\mega \byte} of memory, as well as the optimization necessary to run the pipeline at typical \gls{vio} real-time frame-rates of \SI{20}{\fps} are described in \cref{sec:deployment_optimizations}. We verify the applicability to real-world embedded systems by implementing and benchmarking the \gls{levio} pipeline on a commercially available \gls{soc}. The embedded system runs a self-contained C implementation based on the best-performing Python golden model.

\subsection{The Proposed VIO Pipeline}
\label{sec:levio_pipeline_description}

\begin{figure}[t]
    \centering
    \includegraphics{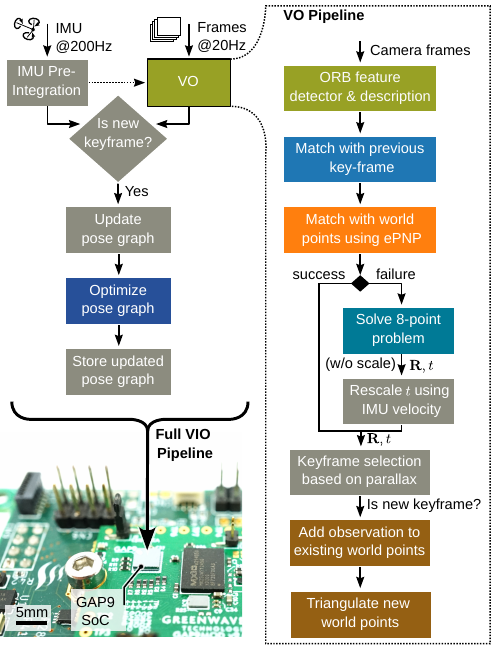}
    \caption{Blockdiagram of the \gls{levio} pipeline. The overview of the full \gls{vio} pipeline is depicted on the left, with a detailed description of the \gls{vo} segments to the right. The complete pipeline has been verified using a Python golden model and is deployed on the GAP9 \gls{soc}.}
    \label{fig:levio-pipeline}
\end{figure}

Following the terminology introduced in \cref{sec:related_work_influential_vio}, the \gls{levio} pipeline, depicted in \Cref{fig:levio-pipeline}, utilizes feature-based (indirect) motion estimation using \gls{orb} features \cite{rublee2011orb}. To estimate relative movements from frame to frame, the 8-point algorithm is used, whereas for estimating the camera pose relative to 3D landmarks, the \gls{epnp} algorithm \cite{lepetit2009ep} is employed. \Gls{levio} uses a pose-graph optimization approach to refine the odometry estimates. To obtain a pose-graph that can be efficiently and accurately optimized, a keyframe selection scheme is employed. To select frames with a sufficiently large baseline for accurate landmark triangulation and pose-graph optimization, keyframes are only introduced once a certain parallax threshold between the current frame and the previous keyframe is exceeded. The pose-graph optimization is formulated as a tightly coupled problem, where \gls{ba} of \gls{vo} estimates and the fusion of pre-integrated \gls{imu} measurements are jointly executed to obtain a \gls{vio} estimate. Due to the limited memory on the \gls{soc} platform, loop closure is omitted, and a monocular camera setup is used.

The characteristics of each pipeline segment, as well as our design decisions, are detailed in the subsequent \cref{sec:vo_segments} and \cref{sec:imu_optimizer_segments}.

\subsection{VO Pipeline Segments}
\label{sec:vo_segments}

The \gls{vo} pipeline depicted on the right side of \Cref{fig:levio-pipeline} is composed of the following segments:
\begin{enumerate}
    \item \textbf{\gls{orb}:} For feature detection and description, we selected \gls{orb} \cite{rublee2011orb}, as it offers a good trade-off between computational efficiency and relevant descriptor properties like re-detectability and uniqueness. The layered approach of efficiently computing the FAST score \cite{rosten2006machine} on each pixel, while only computing the Harris score \cite{harris1988combined} on FAST corners, allows for efficient computation and is well-suited for parallelization. The same is true for the computation of \gls{orb} descriptors, which are only computed on valid Harris corners.
    \item \textbf{Feature Matching:} Feature matching between two frames, as well as between frames and world points, is performed using a two-way brute-force matcher. Given the relatively small number of descriptors (up to 700 for frames and up to 1000 for the world points), this can be done efficiently and is also well parallelizable.
    \item \textbf{8-point \gls{ransac}:} The relative motion between two frames is estimated using the 8-point essential matrix computation \cite{longuet1981computer}.
    The essential matrix, $E$, links normalized point correspondences ($x_i, x_i^{\prime}$) through the epipolar constraint:
    \begin{equation}
    x_{i}^{\top}E~x_{i}^{\prime} = 0
    \label{eq:epipolar}
    \end{equation}
    $E$ is estimated using least-squares, and its rank is constrained to 2. The final rotation ($R$) and translation ($t$) are extracted from the essential matrix within a \gls{ransac} loop \cite{fischler1981random} to obtain an accurate solution in the presence of outliers. The use of \gls{ransac} allows for straightforward parallelization of the essential matrix estimation by distributing \gls{ransac} iterations to different cores. The motion estimation between two frames is performed to bootstrap the \gls{vio} pipeline, i.e., to initialize the pipeline or when not enough feature matches between the current frame and the world points are found.
    \item \textbf{\gls{epnp} \gls{ransac}:} To determine the perspective of the current frame relative to 3D landmarks, matches between image points and world points (landmarks) are needed. The camera's absolute pose ($R, t$) is determined by mapping 3D world landmarks ($X_j$) to their 2D image projections ($x_j$) using the efficient \gls{epnp} algorithm \cite{lepetit2009ep}. \Gls{epnp} solves the 2D-3D correspondence problem by expressing 3D points as a linear combination of four virtual control points. In this case, \gls{ransac} is also used to find an accurate solution in the presence of outliers. Single \gls{ransac} iterations can be distributed to different cores to achieve parallel execution. In the case where not enough matches between world points and the current frame are found (default threshold value of 25), \gls{epnp} \gls{ransac} is not executed, and the motion estimation falls back to 8-point \gls{ransac}.
    \item \textbf{Simple Keyframe Selection:} To reduce the number of frames that are being tracked in the pose-graph, while still keeping enough data for spatio-temporally consistent \gls{ba}, we employ a simple keyframe selection scheme. A frame is selected to be a keyframe and added to the pose-graph once a certain parallax threshold between the current frame and the previous keyframe is exceeded.
    \item \textbf{Triangulation and Tracking of World Points:} Once a new keyframe is added, the correspondences (observations) between the new keyframe and the world points are registered. Furthermore, the feature matches between the current and the previous keyframe are used to triangulate previously unobserved world points.
\end{enumerate}

\subsection{IMU and Optimization Pipeline Segments}
\label{sec:imu_optimizer_segments}

The remaining \gls{vio} segments depicted on the left side of \Cref{fig:levio-pipeline} are required for the processing of \gls{imu} readings and the pose-graph optimization, which simultaneously performs \gls{ba} and jointly optimizes the \gls{imu} constraints and the \gls{vo} constraints. The pipeline segments are the following:

\begin{enumerate}
    \item \textbf{Maintaining the Pose-Graph:} To be able to perform \gls{ba}, which optimizes the pose-graph, we must maintain a pose-graph consisting of poses and world points \cite{macario2022comprehensive}. The observations form the edges between these poses and world points. Furthermore, to account for the \gls{imu} data, each pose contains additional velocity and \gls{imu}-bias states, and the pre-integrated \gls{imu} measurements form the edges between two consecutive poses and their respective states.
    \item \textbf{\gls{imu} Pre-Integration:} To account for both effects of \gls{imu} measurements arriving at different rates than image frames and keyframes being selected at irregular intervals, we employ \gls{imu} pre-integration \cite{forster2016manifold} to obtain only one \gls{imu} constraint between two keyframes.
    Measurements between keyframes $k$ and $k+1$ are pre-integrated into single relative motion constraints. Let $R$, $p$, and $v$ be the rotation, position, and velocity of the body frame, respectively, and $\Delta t$ be the total time interval. $g$ is the gravity vector, and $\Delta\hat{R}$, $\Delta\hat{v}$, and $\Delta\hat{p}$ are the pre-integrated measurements for rotation, velocity, and position. This process yields the following residuals, which are minimized during optimization:
    \begin{equation}
    r_{R} = \log((\Delta\hat{R})^{\top}R_{k}^{\top}R_{k+1})
    \label{eq:res_rot}
    \end{equation}
    \begin{equation}
    r_{v} = R_{k}^{\top}(v_{k+1} - v_{k} - g\Delta t) - \Delta\hat{v}
    \label{eq:res_vel}
    \end{equation}
    \begin{equation}
    r_{p} = R_{k}^{\top}(p_{k+1} - p_{k} - v_{k}\Delta t - \frac{1}{2}g{\Delta t}^{2}) - \Delta\hat{p}
    \label{eq:res_pos}
    \end{equation}
    \item \textbf{Pose-Graph Optimization} The \gls{ba} of the \gls{vo} estimates and the fusion with the pre-integrated \gls{imu} data is formulated as a tightly coupled optimization problem. Whenever a new keyframe is added, we stop the current \gls{imu} pre-integration and add both the new observation and \gls{imu} constraints to the optimization problem. The optimization problem is formulated as a moving window on the pose-graph, only optimizing the most recent keyframes. The oldest keyframe states (pose, velocity, and bias) serve as initial conditions and are further anchored by additional constraints.
\end{enumerate}

\subsection{Deployment Optimizations}
\label{sec:deployment_optimizations}
Moving from the Python-based golden model to ultra-low-power embedded platforms, the following functional additions and changes were necessary to run \gls{levio}. To verify the applicability of \gls{levio} to ultra-low-power embedded systems, we target the specifications of the commercially available GAP9 \gls{soc}. GAP9 hosts a nine-core compute cluster comprising a master core and eight worker cores, as well as an additional fabric controller core responsible for interacting with peripherals and external memories. Operating at its maximum frequency of \SI{370}{\mega \hertz} GAP9 uses less than \SI{100}{\milli \watt} \cite{kiamarzi2024qr,cereda2024device}.

\begin{enumerate}
    \item \textbf{Custom Linear Algebra Library:} To our knowledge, there is no linear algebra library available in C that supports advanced operations like \gls{svd} or the Jacobi Eigenvalue algorithm, which is compact enough to be deployed on an \gls{soc} like GAP9. Established feature-rich libraries like LAPACK \cite{anderson1999lapack} could not be run on GAP9. Therefore, we built a custom linear algebra library that supports \gls{svd} of square matrices, the Jacobi Eigenvalue algorithm, the Inverse Power method, and a selection of solvers for systems of linear equations.
    \item \textbf{Hierarchical Memory Handler:} The memory hierarchy of GAP9 consists of \SI{128}{\kilo \byte} of L1 memory, which is accessible by the compute cluster and \SI{1.6}{\mega \byte} of L2 memory only accessible by the fabric controller. While the \SI{128}{\kilo \byte} is sufficient to store the required values for the function or pipeline segment being executed at any given time, it is not enough to store all relevant data, most notably the pose-graph and the data of the previous keyframe. Therefore, we allocate designated space for this data in the L2 memory and move it into the L1 memory when needed. The L1 memory is mainly used as a scratchpad, and no data is preserved between iterations.
    \item \textbf{Reduced Image Resolution:} Given the \SI{128}{\kilo\byte} limitation of the L1 memory, we decided to use QQVGA (160 pixels by 120 pixels) resolution for the image inputs, to have sufficient work memory available for the online computations.
    \item \textbf{Adapted Optimizer Formulation:} 
    The \gls{vio} problem is formulated as a \gls{map} estimation solved with the \gls{lm} algorithm. To make the optimization problem tractable on a \gls{soc}, we exploit the problem's sparse structure using the Schur complement \cite{schur1918potenzreihen} on the full system of normal equations. The full Hessian matrix $H$ and the corresponding right-hand side vector $b$ are partitioned based on the two types of states: the camera pose states ($p_p$) and the 3D landmark states ($p_l$), with $\delta p$ being the perturbation vector:
    \begin{equation}
    \begin{bmatrix} H_{pp} & H_{pl} \\ H_{lp} & H_{ll} \end{bmatrix} \begin{bmatrix} \delta p_p \\ \delta p_l \end{bmatrix} = \begin{bmatrix} b_p \\ b_l \end{bmatrix}
    \label{eq:full_hessian}
    \end{equation}
    This marginalizes the 3D landmark states ($p_l$), resulting in a much smaller, reduced system that solves only for the camera pose states ($p_p$):
    \begin{equation}
    (H_{pp} - H_{pl}H_{ll}^{-1}H_{lp}) \delta p_p = b_p - H_{pl}H_{ll}^{-1}b_{l}
    \label{eq:schur}
    \end{equation}
    Solving this condensed system drastically reduces computational and memory overhead.
    Furthermore, due to the high timing requirements for \gls{lm} optimization, we do not sequentially execute the optimizer after a new keyframe has been added, but we continuously run the optimization on the fabric controller, largely independent of the \gls{vo} iterations. However, the optimization problem is updated whenever a new keyframe is added.
    \item \textbf{Parallelization:} We adapted most \gls{vo} pipeline segments to be executable in a parallelized fashion. The \gls{orb} feature detection and description, as well as the brute-force matching logic, have been parallelized by processing subsets of the input data on different cores. Furthermore, both the 8-point and the \gls{epnp} \gls{ransac} loops have been parallelized by processing single \gls{ransac} iterations on different cores.
\end{enumerate}

\begin{figure}[t]
    \centering
    \includegraphics{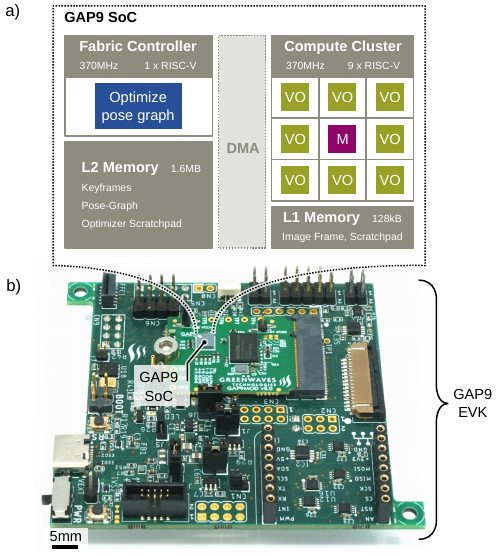}
    \caption{a) Block diagram presenting the hardware resources of the GAP9 \gls{soc} and tasks of the \gls{levio} pipeline, with the pose graph optimization running on the fabric controller and the \gls{vo} mapped to the eight parallel processing cores. Non-parallelized parts of the \gls{vo} are handled by the Master Core (M). b) Overview of the physical GAP9 Evaluation Kit (EVK).}
    \label{fig:levio-HW-Mapping}
\end{figure}

\section{Evaluation}

\label{sec:levio_evaluation}
The evaluation of the \gls{levio} pipeline was performed both at an algorithmic level utilizing the Python golden model, and by analyzing its viability and employability on real-world embedded platforms by deploying the optimized C implementation on the GAP9 evaluation kit, as shown in \Cref{fig:levio-HW-Mapping}.

\subsection{Performance Evaluation of the LEVIO Algorithm}
To assess the algorithmic capabilities of the \gls{levio} system, we rely on the golden model as mentioned in \cref{sec:levio_pipeline_description}, developed to estimate how the addition or removal of certain pipeline segments affects the overall prediction accuracy while also considering the computational requirements. 

Therefore, we benchmark multiple pipeline configurations using the \emph{EuRoC} dataset \cite{burri2016euroc}. Furthermore, we performed an extensive parameter sweep on relevant \gls{vio} parameters to further increase the statistical relevance of the design space exploration. For each of the pipeline variants and parameter choices, the resulting trajectory estimate was evaluated using the \emph{rpg\_trajectory\_evaluation}\footnote{\url{https://github.com/uzh-rpg/rpg_trajectory_evaluation}} tool \cite{zhang2018tutorial}. We focus our analysis on the absolute \gls{rmse}, i.e., the \gls{rmse} over all predicted poses, as well as the relative translation error for multiple trajectory segments of \SI{40}{\meter}, covering the full length of each trajectory. To obtain a single metric, we state the average error across all sampled segments.

\subsection{Real World Embedded Evaluation of LEVIO}
To verify that the \gls{levio} pipeline is viable for use on real-world resource-constrained systems, we deploy it on the GAP9 parallel ultra-low-power \gls{soc} by GreenWaves Technologies. While the deployed pipeline is conceptually the same as the golden model, some modifications and additions, as stated in \cref{sec:deployment_optimizations}, were necessary.

For the deployed pipeline on GAP9, we analyse the computational requirements in terms of cycle counts when executing \gls{levio} on a single core and when parallelizing the execution on the available eight worker cores. Additionally, we assess the achieved parallelization speedup for each of the parallelized code segments, as well as the overall performance.

Furthermore, we analyze the utilization of both the L1 and L2 memories. As we store the pose-graph persistently in L2 memory, this is a static number. On the contrary, for L1 and the L2 memory used as work memory by the pose-graph optimizer, we analyze the peak utilization for each pipeline segment.

\subsection{Benchmark Dataset: EuRoC}
For all system evaluations, we used the \emph{EuRoC} dataset \cite{burri2016euroc}.
This dataset is a challenging standard in the \gls{vio} community, recorded on board a real drone under high-dynamic and variable-illumination conditions, making it the most application-relevant benchmark for the \gls{levio} system.
While \emph{EuRoC} was recorded with a drone that is larger than the systems considered in this work, the used sensors are also representative of resource-constrained devices as considered in this work. \emph{EuRoC} offers global-shutter stereo camera data at 20 \gls{fps} and \gls{imu} readings at \SI{200}{\hertz}. Since those are typical data rates for \gls{vio} systems, we use them both in our golden model and the GAP9 deployment. We do, however, only use one camera, as \gls{levio} is a monocular \gls{vio} system. Additionally, we subsample the images when deploying the pipeline on GAP9.

\begin{figure}[t]
    \centering
    \includegraphics{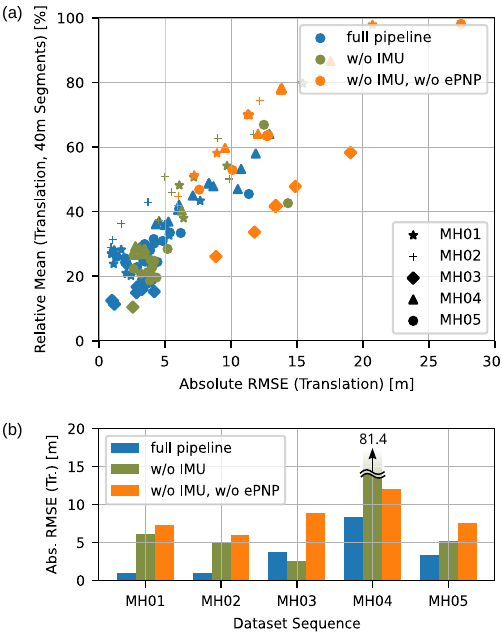}
    \caption{Relative and absolute translation RMS error of the golden model during the parameter sweep (a) and absolute RMSE error for the best performing configuration, when applied to each sub-dataset (b). Data is color-coded for the type of pipeline, and the shapes indicate the dataset it was evaluated on (EuRoC MH01-MH05).}
    \label{fig:rmse_rel_error_parametersweep}
\end{figure}

\begin{table*}[t]
    \centering
    \caption{The performance numbers of the best performing parameter set for each pipeline configuration in terms of \gls{rmse}, when averaging over the five \emph{EuRoC} sequences. Additionally, for the full \gls{levio} pipeline, the performance numbers of the best-performing parameter set in terms of translation error are given. Whereas for the other two pipeline configurations, the best performing parameter sets in terms of \gls{rmse} and relative translation error are identical.}
    \begin{tabular}{l|rr|rr|rr|rr|rr}
        \toprule 
         & \multicolumn{2}{c|}{MH01} & \multicolumn{2}{c|}{MH02} & \multicolumn{2}{c|}{MH03} & \multicolumn{2}{c|}{MH04} & \multicolumn{2}{c}{MH05} \\
        \textbf{Pipeline Configuration} & Abs. [m] & Rel. [\%] & Abs. [m] & Rel. [\%] & Abs. [m] & Rel. [\%] & Abs. [m] & Rel. [\%] & Abs. [m] & Rel. [\%]   \\
        \midrule 
        \gls{levio} w/o \gls{imu}, w/o \gls{epnp} 
        &  7.228 & 51.23
        &  5.982 & 41.77
        &  8.867 & 26.10
        & 12.061 & 64.10
        &  7.601 & 46.80 \\

        \gls{levio} w/o \gls{imu}
        &  6.092 &  48.14
        &  4.995 &  50.77
        &  \textbf{2.572} &  \textbf{10.42}
        & 81.374 & 282.09
        &  5.202 &  28.45 \\

        Full \gls{levio} - Best Abs. Error
        & \textbf{0.963} &  \textbf{26.96}
        & \textbf{0.920} &  \textbf{28.86}
        & 3.685 &  18.61
        & \textbf{8.334} &  \textbf{48.77}
        & \textbf{3.376} &  \textbf{28.29} \\
        \midrule

        Full \gls{levio} - Best Rel. Error
        &  1.115 & 23.83
        &  4.431 & 19.36
        &  2.843 & 14.89
        & 10.781 & 57.98
        &  3.467 & 29.96 \\
        \bottomrule
    \end{tabular}
    \label{tab:best_performing_parameter_sets}
\end{table*}

\section{Results}
\label{sec:levio_results}
In this section, we present the results obtained from the algorithmic study performed using the golden model and the results and insights gained when deploying \gls{levio} on GAP9. For the algorithmic study, we performed an ablation study of varying \gls{vio} pipeline complexities to motivate the use of pipeline segments, as well as the parameter selection. For the deployment on GAP9, we report relevant metrics for both real-time robotic applications and resource usage, such as latency, memory usage, and parallelization performance.

\subsection{Pipeline Evaluation}
We analyzed three pipeline configurations as described below:

\begin{enumerate}
    \item \textbf{Full \gls{levio} Pipeline:} The full pipeline consists of all pipeline segments as described in \cref{sec:levio_methodology} and shown in \Cref{fig:levio-pipeline}.
    \item \textbf{\gls{levio} without \gls{imu} Data:} For the second configuration we reduced \gls{levio} to a \gls{vo}-only configuration, where no \gls{imu} data is included in the optimization step.
    \item \textbf{\gls{levio} without \gls{imu} Data and \gls{epnp}:} In this configuration, we reduce the system to the 8-point algorithm for frame-to-frame pose estimation and use keyframe selection and \gls{ba} to optimize the result.
\end{enumerate}

Furthermore, we used parameter sweeps across the following dimensions: 
\begin{enumerate*}
    \item Optimization Window Size,
    \item Keyframe Threshold,
    \item Reprojection Error Noise Model,
    \item IMU Noise Model, and
    \item Gravity Vector Updates,
\end{enumerate*}
where the last two parameters are only relevant for the full pipeline, where \gls{imu} data is being used.

The accuracy results of those benchmark sweeps are depicted in \Cref{fig:rmse_rel_error_parametersweep}. Furthermore, \Cref{fig:rmse_rel_error_parametersweep}b) indicates the best-performing parameter set for each pipeline configuration in terms of \gls{rmse}. In \Cref{tab:best_performing_parameter_sets}, we list the performance figures of those three configurations in terms of \gls{rmse}. Additionally, we list the best-performing models in terms of relative translation error. For the full \gls{levio} pipeline, the best relative translation error is achieved by a different parameterization than the best \gls{rmse}.

One can observe that the most complex pipeline, Full \gls{levio}, reaches the highest accuracy when adequately parametrized, with an average \gls{rmse} of \SI{3.46}{\meter} over all sequences. Additionally, \Cref{fig:rmse_rel_error_parametersweep}a) illustrates that the addition of \gls{imu} data mainly reduces long-term drift. This can be observed by the relative mean translation error of \SI{40}{\meter} sub-trajectories being in a similar range for the full pipeline and the pipeline without \gls{imu} data, but the \gls{rmse}, which captures the full trajectory, being lower for the full pipeline. When looking at the data of the best performing configurations in \Cref{tab:best_performing_parameter_sets} for MH05, we can observe that the relative error for both pipelines is nearly identical. In contrast, the absolute \gls{rmse} of the pipeline without \gls{imu} data is with \SI{5.2}{\meter} versus \SI{3.38}{\meter} approximately \SI{54}{\%} higher than for the full pipeline.

In \Cref{fig:rmse_rel_error_parametersweep}a), we can observe that depending on the parameterization, there is a trade-off between \gls{rmse} and relative translation error. The scatter plot indicates that the \gls{rmse} can be increased to, in turn, reduce the relative translation error. Depending on the application and whether short-term or long-term consistency is more important, this can be an interesting tuning parameter.

\subsection{Deployment on GAP9}
\begin{figure}[t]
    \centering
    \includegraphics{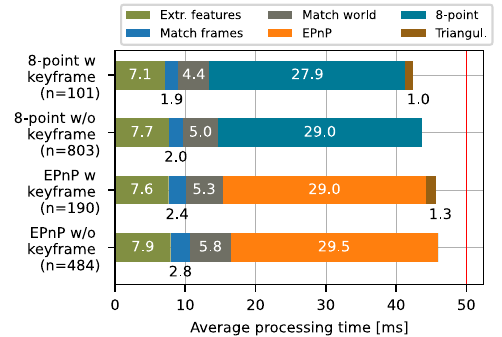}
    \caption{Average runtime of the parallelized LEVIO pipeline for different scenarios occurring during processing (either \gls{epnp} or 8-point, and with/without keyframe). Small numbers showing the execution time in ms. On average, the pipeline processes incoming camera frames with a runtime below \SI{50}{\milli\second}, allowing for \SI{20}{\fps}.}
    \label{fig:gap9_pipeline_timing}
\end{figure}

To evaluate the real-time capabilities of \gls{levio} on GAP9, we analyzed the timing requirements for the \gls{vo} part of the pipeline in \Cref{fig:gap9_pipeline_timing},  with the computation limit imposed by a \SI{20}{\fps} image stream indicated by the red line. To obtain statistical runtime data, we process the MH01 sequence on GAP9 (first 1578 frames). Using this real-world dataset yields a varying number of matches between the triangulated landmarks and the image frames or between two image frames. Given the varying number of matches, both the 8-point and the \gls{epnp} algorithm are executed regularly, in \SI{57.3}{\%} and \SI{42.7}{\%} of the \gls{vo} iterations, respectively.

\begin{table}[t]
    \begin{center}
    \caption{Breakdown of average compute cycles needed when processing the \emph{EuRoC} dataset, in case of single core and parallelized multicore execution on the GAP9 platform}.
    \label{tab:compute_breakdown}
    \begin{tabular}{l|rrr}
    \toprule
     & \multicolumn{1}{c}{\textbf{Single-Core}}  & \multicolumn{1}{c}{\textbf{Multi-Core}} & \multicolumn{1}{c}{\textbf{Speed Up}} \\
    \textbf{Pipeline Step}& {[kCycles]} $\downarrow$ & {[kCycles]} $\downarrow$ & [Factor] $\uparrow$ \\
    \midrule
    Feature Extraction & 21817 & 2926 & 7.46 \\
    Frame Matching & 8112 & 1033 & 7.85 \\
    World Matching & 16782 & 2128 & 7.89\\
    \midrule
    \gls{epnp}  \gls{ransac}$^a$ & 57324 & 10899 & 5.26\\
    8-point  \gls{ransac}$^a$ & 50391 & 10323 & 4.88\\
    \midrule
    Triangulation$^b$ & 374 & 374 & 1.00\\
    \bottomrule
    \multicolumn{4}{l}{}\\[-0.8em]
    \multicolumn{4}{l}{$^a$\textit{either \gls{epnp} or 8-point are calculated}}\\
    \multicolumn{4}{l}{$^b$\textit{only done for keyframes, not parallelized}}\\
    \end{tabular}
    \end{center}
\end{table}

As can be seen from the timing analysis shown in \Cref{fig:gap9_pipeline_timing}, the complexity of the full \gls{levio} pipeline over the configurations without \gls{imu} data and \gls{epnp} computation comes at a minimal latency overhead. To a great extent, this is due to the mutually exclusive execution of the \gls{epnp} and the 8-point algorithms. We only start \gls{epnp} when we have sufficiently many feature matches with landmarks; otherwise, the 8-point algorithm is executed. Both 8-point and \gls{epnp} \gls{ransac} are computationally heavy, each utilizing between \SI{62.9}{\%} and \SI{64.1}{\%} of the computation time per iteration. Secondly, the overhead of the \gls{imu} pre-integration is minimal, contributing only 2.42\,kCycles per \gls{imu} measurement, or 24.2\,kCycles to pre-integrate the ten measurements obtained between two \gls{vo} iterations, introducing an overhead of less than \SI{0.15}{\%}. Furthermore, the \gls{imu} pre-integration can be performed on either the cluster or the fabric controller.

The resulting parallelization speedup when running the \gls{vo} pipeline on the eight worker cores of GAP9 versus the execution on a single core is shown in \Cref{tab:compute_breakdown}. This comparison effectively isolates the algorithmic contributions from the hardware capabilities, demonstrating the performance advantages gained by fully utilizing \gls{levio}'s parallelization capability on the hardware platform. Furthermore, this architecture allows for highly efficient operating-point tuning: Since the processing throughput scales linearly with the clock frequency, the clock speed can be reduced linearly if a lower update rate (e.g., 10 FPS) is sufficient for the application. This capability enables the system to minimize dynamic power consumption while strictly adhering to the required real-time constraints.

For the feature extraction and both feature matching implementations (frame and world matching), the parallelization is very effective. In those cases, the parallelization is achieved by splitting the input data as evenly as possible across the eight worker cores. Additionally, the feature extraction and matching algorithms have a nearly constant runtime complexity. For both \gls{ransac} algorithms, the parallelization is less effective. This is due to the fact that the parallelization is achieved by distributing the \gls{ransac} iterations across the worker cores, while the runtime required for a single \gls{ransac} iteration can vary significantly depending on the numerical conditioning of the randomly selected input samples. The runtime differences are caused by the underlying Jacobi Eigenvalue computation used for the \gls{svd}, which is an iterative algorithm whose convergence rate is strongly dependent on the input data. Thus, during each compute batch, the compute cluster needs to wait for all \glspl{svd} to converge before continuing. Furthermore, given the small computational cost, the triangulation of landmarks has not been parallelized.

The pose-graph optimization is computationally expensive at up to 55\,MCycles or \SI{15}{\milli \second} per iteration, but can be performed independently of the \gls{vo} estimation. Therefore, it is executed on the single fabric controller core, performing three optimizer iterations per \gls{vo} iteration. Once a new keyframe is selected, the optimizer constraints are updated.

Regarding the memory utilization, we analyze the peak utilization for each pipeline segment in \Cref{tab:memory_breakdown}. Since we use the L1 memory as a scratchpad and only preserve the minimally necessary data between the pipeline segments, one can observe fluctuating memory occupation, but also a trend that an increasing amount of L1 memory is being used throughout a single \gls{vo} iteration. At the end of the \gls{vo} iteration, all data is transferred into the persistent L2 blocks storing the most recent keyframe and the pose-graph consisting of the eight most recent keyframe poses, the thousand most recent landmarks, and the corresponding observations linking the keyframe poses to the landmarks. To construct the full optimization problem, consisting of the Jacobian terms of each observation, the optimizer requires up to \SI{743.6}{\kilo \byte} of L2 memory.

\begin{table}[t]
    \centering
    \caption{Memory Usage of various pipeline segments of \gls{levio}. L1 and L2 indicate in which memory subsystem the data is stored. For the pipeline segments that use more work memory, when executed in parallel, we indicate the respective usage next to the single-core usage.}
    \begin{tabular}{l|c|rr|rr}
    \toprule
     & \textbf{Memory}  & \multicolumn{2}{c|}{\textbf{Usage} [kbyte] $\downarrow$} & \multicolumn{2}{c}{\textbf{Occupation} [\%] $\downarrow$} \\
    \textbf{Pipeline Step} &  & Single & Parallel & Single & Parallel\\
    \midrule
    Feature Extraction  & L1 & 64.0 & & 50.03 & \\
    World Matching & L1 & 78.7 & & 61.52 & \\
    \gls{epnp} \gls{ransac} & L1 & 82.2 & 103.8 & 64.23 & 81.12  \\
    Frame Matching & L1 & 61.3 & & 47.88 & \\
    8-point \gls{ransac} & L1 & 65.6 & 69.2 & 51.26 & 54.08 \\
    Triangulation & L1 & 109.1 & & 85.23 & \\
    \midrule
    Last Keyframe & L2 & 50.9 & & 3.18 & \\
    Pose-Graph & L2 & 64.2 & & 4.01 & \\
    Optimizer & L2 & 743.6 & & 46.47 & \\
    \bottomrule
    \end{tabular}
    \label{tab:memory_breakdown}
\end{table}

\section{Discussion}
\label{sec:levio_discussion}

This work demonstrates that full six \gls{dof} \gls{vio} can be achieved in real time on highly constrained embedded systems with memories in the \si{\mega \byte}-range and compute power in the range of giga-operations per second. \Gls{levio} leverages visual tracking (\gls{orb}, \gls{epnp}) and inertial fusion techniques (\gls{imu} pre-integration, pose-graph optimization), while introducing a hardware-aware \gls{vio} pipeline optimized for parallel execution and low memory usage.

On the GAP9 \gls{soc}, \gls{levio} enables real-time performance at \SI{20}{\fps}, benefiting from parallelized \gls{ransac} loops, efficient memory management, and a lightweight custom optimizer for \gls{ba}. While framerates of \SI{20}{\hertz} are typical and sufficient for most \gls{vio} tasks \cite{burri2016euroc}, the \gls{levio} pipeline is suitable for higher framerates, when deployed on corresponding hardware. 

In terms of efficiency, highly power-efficient \gls{vio} pipelines can be realized using specialized \glspl{asic} (e.g. Navion \cite{suleiman2019navion}). However, the development is expensive and offers little to no reconfigurability to adapt to algorithmic advances. Being implemented on a general-purpose multicore \gls{soc}, the presented \gls{levio} implementation strikes a strong balance between software specialization and reconfigurability. Thus, \gls{levio} stands out for providing highly efficient full six-\gls{dof} \gls{vio} estimation without requiring \glspl{asic}, \glspl{gpu}, or simplified motion models.

Furthermore, the results demonstrate that classical single-core microcontrollers are currently incapable of executing a fully featured \gls{vio} pipeline in real-time (requiring roughly \SI{300}{\milli \second} per iteration at \SI{370}{\mega \hertz}). \Gls{levio} highlights the growing potential of lightweight multicore platforms to serve as a practical and scalable foundation for robotics and autonomous systems operating under tight resource constraints.

\subsection{Comparison with Influential VIO Systems}
We benchmark \gls{levio} against VINS-Mono \cite{qin2018vins} and ORB-SLAM3 \cite{campos2021orb} on the EuRoC dataset (see \cref{tab:ate_comparison}), evaluating the baselines at both standard (WVGA) and downscaled (QVGA) resolutions. A fundamental distinction lies in the algorithmic scope: while the baselines rely on Loop Closure (LC) to correct global drift, achieving sub-meter accuracy ($<0.3$ m), \gls{levio} functions as a pure \gls{vio} pipeline, prioritizing constant-time processing over global consistency. The results expose the fragility of desktop-class algorithms under embedded constraints, with ORB-SLAM3 failing catastrophically on MH\_04 at QVGA resolution. Conversely, \gls{levio} trades absolute accuracy (yielding $0.9$--$3.5$ m error) for strict robustness within a $100$ mW power envelope. This represents a projected $15\times$ efficiency gain over VINS-Mono running on systems like the Jetson Nano, confirming \gls{levio} as a viable solution for micro-drone stabilization where low-latency availability outweighs the need for global mapping.

\begin{table}[t]
\centering
\caption{Absolute Trajectory Error (ATE) RMSE [m] on EuRoC MAV sequences (Machine Hall).}
\label{tab:ate_comparison}
\setlength{\tabcolsep}{3.1pt}
\begin{tabular}{@{}lccccccc@{}}
\toprule
\textbf{Method} & \textbf{Res.} & \textbf{LC} & \multicolumn{5}{c}{\textbf{Sequence (ATE RMSE [m])}} \\
\cmidrule(lr){4-8}
& & & MH\_01 & MH\_02 & MH\_03 & MH\_04 & MH\_05 \\
\midrule
LEVIO & Std. & No & 0.96 & 0.92 & 3.69 & 8.33 & 3.38 \\
LEVIO & QVGA & No & 1.82 & 1.44 & 2.10 & 7.49 & 4.48 \\
\midrule
ORB-SLAM3 & Std. & Yes & 0.15 & 0.18 & 0.27 & 0.37 & 0.27 \\
ORB-SLAM3 & QVGA & Yes & 0.18 & 0.22 & 0.26 & 1528.13 & 0.93 \\
\midrule
VINS-Mono & Std. & No & 0.20 & 0.23 & 0.32 & 0.47 & 0.41 \\
VINS-Mono & QVGA & No & 0.25 & 0.18 & 0.46 & 0.47 & 0.60 \\
\midrule
VINS-Mono & Std. & Yes & 0.14 & 0.16 & 0.30 & 0.34 & 0.34 \\
VINS-Mono & QVGA & Yes & 0.25 & 0.18 & 0.40 & 0.41 & 0.54 \\
\bottomrule
\end{tabular}
\end{table}

\subsection{Comparison with Ultra-Low-Power VIO}
The landscape of ultra-low-power \gls{vio} and \gls{vo} systems presents a trade-off between power consumption, computational complexity, and full 6 \gls{dof} capability, as shown in \Cref{tab:embedded_vio_refs}. \Gls{levio} occupies a unique niche by offering a full 6 \gls{dof} \gls{vio} pipeline with advanced features within the sub-\SI{100}{\milli\watt} range on \gls{cots} hardware. A comparison with low-power approaches highlights this differentiation: 
\begin{itemize} 
    \item \textit{Navion \cite{suleiman2019navion}} achieves ultra-low power (\SI{2}{\milli \watt}) utilizing a custom \gls{asic}, which is less accessible and flexible than \gls{levio}'s \gls{cots} platform.
    \item \textit{PX4FLOW \& Parallelized PX4FLOW \cite{honegger2013open, kuhne2022parallelizing}} operate at $<$\SI{100}{\milli\watt} but are limited to simplified 2D planar motion using basic optical flow, unlike \gls{levio}'s full 6 DoF capability.
    \item \textit{PicoVO \cite{he2021picovo}} is a 3D Monocular \gls{vo} system requiring higher power (\SI{310}{\milli\watt}). \Gls{levio} runs at a significantly lower power point and includes inertial fusion (\gls{vio}), which enhances accuracy compared to \gls{vo}-only methods. 
\end{itemize}
In essence, \gls{levio} resolves the trade-off by delivering high algorithmic complexity on an accessible, ultra-low-power \gls{cots} platform, avoiding the compromises of custom hardware, power requirements, or algorithmic simplification common in similar embedded systems.

\section{Conclusion}
\label{sec:levio_conclusion}

We presented \gls{levio}, a lightweight and fully embedded \gls{vio} pipeline optimized for ultra-low-power platforms with power envelopes below \SI{100}{\milli \watt}. \Gls{levio} achieves robust six \gls{dof} tracking while operating within tight computational and memory budgets (using less than \SI{110}{\kilo \byte} L1 and $\approx$\SI{850}{\kilo \byte} L2 memory), offering a compelling balance between efficiency and accuracy. While it does not aim to compete with high-end \gls{vio} systems, \gls{levio} demonstrates that practical, infrastructure-free localization is achievable on resource-constrained devices. This work further demonstrates that parallelization is a valid approach to enable real-time \gls{vio} at \SI{20}{\fps}, suitable for nano drones and smart glasses. Validated on the public \emph{EuRoC} dataset and deployed on GAP9, it enables real-time performance on microcontroller and ultra-low-power \gls{soc} hardware. Additionally, it is available as open-source for further research. 

While \gls{levio} achieves impressive efficiency by successfully implementing a full \gls{vio} pipeline on an ultra-low-power \gls{soc}, the current system has several limitations that guide future research.
To meet the strict power and memory budgets, \gls{levio} currently relies on very low-resolution (QQVGA) images and forgoes loop closure, which may limit long-term global consistency in large-scale environments. Future research will investigate the following open challenges: 
\begin{enumerate*}
    \item loop-closure under tight memory constraints,
    \item resolution–accuracy scaling beyond QQVGA inputs, and
    \item the design of hardware accelerators for typical \gls{vio} operations, such as \gls{svd}, feature extraction, or sparse matrix solvers.
\end{enumerate*}
Additionally, we aim to extend \gls{levio}'s validation to more diverse real-world scenarios and to investigate portability to other low-power embedded architectures.

\bibliographystyle{IEEEtran}
\bibliography{main}

\begin{IEEEbiography}[{\includegraphics[width=1in,height=1.25in,clip,keepaspectratio]{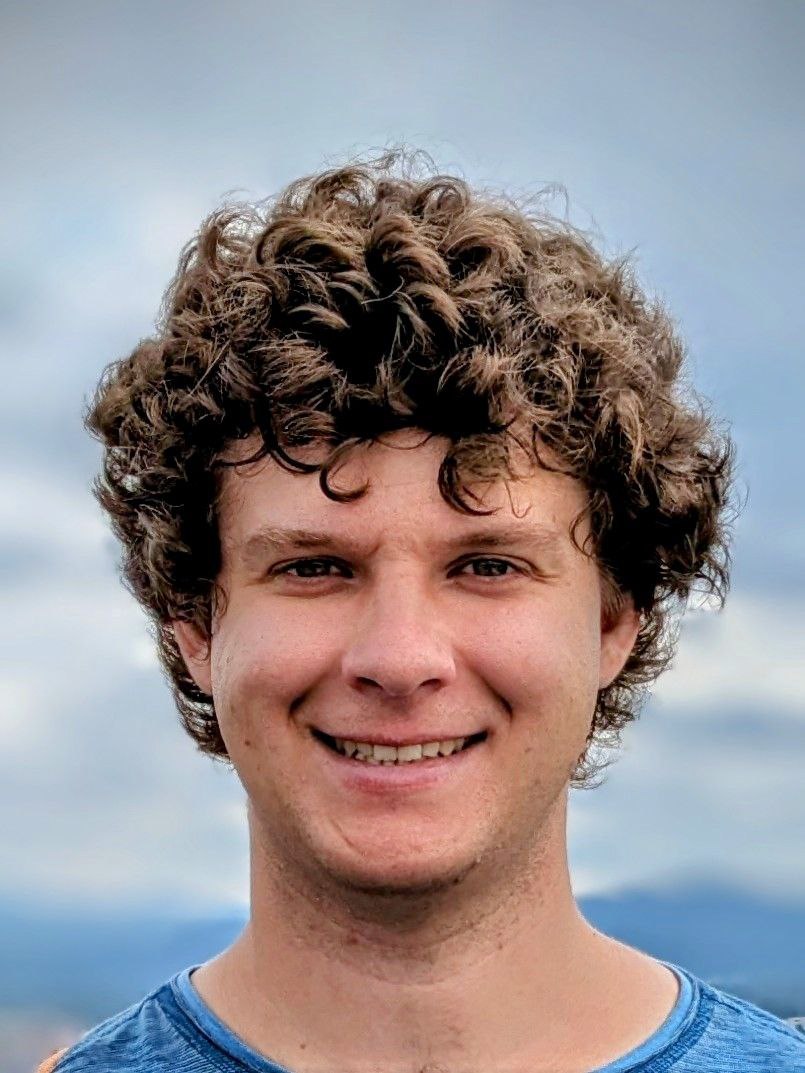}}]{Jonas Kühne}
(Member, IEEE) received the B.Sc. and M.Sc. degrees in electrical engineering and information technology from ETH Zürich, Zürich, Switzerland, in 2016 and 2018, respectively. Between 2019 and 2021, he worked for Agtatec AG, which is part of Assa Abloy. \\
He is currently pursuing his Ph.D. degree with both the Integrated Systems Laboratory and the D-ITET Center for Project-Based Learning at ETH Zürich, Zürich, Switzerland. \\ 
His research interests include algorithm and hardware design for visual inertial odometry and SLAM on low-power embedded systems. 
\end{IEEEbiography}

\begin{IEEEbiography}[{\includegraphics[width=1in,height=1.25in,clip,keepaspectratio]{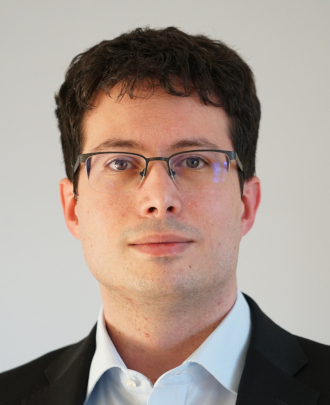}}]{Christian Vogt} (Member, IEEE) received the M.Sc. degree and the Ph.D. in electrical engineering and information technology from ETH Zürich, Zürich, Switzerland, in 2013 and 2017, respectively. He is currently a post-doctoral researcher and lecturer at ETH Zürich, Zürich, Switzerland. His research work focuses on signal processing for low power applications, including field programmable gate arrays (FPGAs), IoT, wearables and autonomous unmanned vehicles.
\end{IEEEbiography}

\begin{IEEEbiography}[{\includegraphics[width=1in,height=1.25in,clip,keepaspectratio]{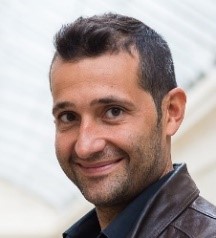}}]{Michele Magno}
(Senior Member, IEEE) received his master's and Ph.D. degrees in electronic engineering from the University of Bologna, Bologna, Italy, in 2004 and 2010, respectively. \\
Currently, he is a \emph{Privatdozent} at ETH Zurich, Zurich, Switzerland, where he is the Head of the Project-Based Learning Center. He has collaborated with several universities and research centers, such as Mid University Sweden, where he is a Guest Full Professor. He has published more than 150 articles in international journals and conferences, in which he got multiple best paper and best poster awards. The key topics of his research are wireless sensor networks, wearable devices, machine learning at the edge, energy harvesting, power management techniques, and extended lifetime of battery-operated devices.
\end{IEEEbiography}

\begin{IEEEbiography}[{\includegraphics[width=1in,height=1.25in,clip,keepaspectratio]{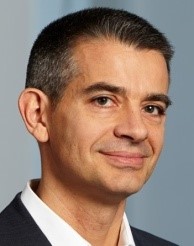}}]{Luca Benini}
(Fellow, IEEE) received the Ph.D. degree in electrical engineering from Stanford University, Stanford, CA, USA, in 1997.\\
He holds the Chair of Digital Circuits and Systems at ETH Zurich, Zurich, Switzerland, and is a Full Professor at the University of Bologna, Bologna, Italy. His current research interests include energy-efficient computing systems' design from embedded to high performance.\\
Dr. Benini is a fellow of the ACM and a member of the Academia Europaea. He was a recipient of the 2016 IEEE CAS Mac Van Valkenburg Award and the 2023 McCluskey  Award.
\end{IEEEbiography}

\end{document}

%% file: acr.tex
\newacronym{vio}{VIO}{Visual Inertial Odometry}
\newacronym{vo}{VO}{Visual Odometry}
\newacronym{imu}{IMU}{Inertial Measurement Unit}
\newacronym{slam}{SLAM}{Simultaneous Localization and Mapping}
\newacronym{levio}{\emph{LEVIO}}{\emph{Lightweight Embedded Visual Inertial Odometry}}
\newacronym{cots}{COTS}{Commercial Off-The-Shelf}
\newacronym{dof}{DoF}{Degrees of Freedom}
\newacronym{soc}{SoC}{System on a Chip}
\newacronym{asic}{ASIC}{Application Specific Integrated Circuit}
\newacronym{ba}{BA}{Bundle Adjustment}
\newacronym{fpga}{FPGA}{Field Programmable Gate Array}
\newacronym{epnp}{EPnP}{Efficient Perspective-\emph{n}-Point}
\newacronym{ransac}{RANSAC}{RANdom SAmple Consensus}
\newacronym{svd}{SVD}{Singular Value Decomposition}
\newacronym{fps}{FPS}{Frames Per Second}
\newacronym{of}{OF}{Optical Flow}
\newacronym{ekf}{EKF}{Extended Kalman Filter}
\newacronym{lm}{LM}{Levenberg–Marquardt}
\newacronym{rmse}{RMSE}{Root Mean Square Error}
\newacronym{vislam}{VI-SLAM}{Visual-Inertial SLAM}
\newacronym{os}{OS}{Operating System}
\newacronym{orb}{ORB}{Oriented FAST and Rotated BRIEF}
\newacronym{gpu}{GPU}{Graphics Processing Unit}
\newacronym{map}{MAP}{Maximum a Posteriori}